\documentclass{INTERSPEECH2023}

\usepackage[table]{xcolor}
\usepackage[T1]{fontenc}
\usepackage{booktabs}
\usepackage{amsmath}

\newcommand{\tmmlp}{\operatorname{TM-MLP}}

\interspeechcameraready

\usepackage{graphicx}
\usepackage{caption}
\usepackage{subcaption}

\definecolor{Gray}{gray}{0.9}

\title{HyperConformer: Multi-head HyperMixer for Efficient Speech Recognition}

\name{
Florian Mai $^{\star,\dagger,\ddagger}$, Juan Zuluaga-Gomez $^{\star,\dagger,\ddagger}$, Titouan Parcollet $^{\mathparagraph}$, Petr Motlicek $^{\dagger,\mathsection}$
\thanks{
$^{\star}$Equal contribution. Order is determined by a coin flip.\\
Research supported by the Swiss National
Science Foundation (\textit{LAOS}, grant 200021-178862) and EU-H2020 (\textit{CRiTERIA}, grant 101021866).
}
}
\address{
  $^{\dagger}$ Idiap Research Institute, Martigny, Switzerland \\
  $^{\ddagger}$ Ecole Polytechnique Federale de Lausanne (EPFL), Switzerland \\
  $^{\mathparagraph}$ University of Cambridge, United Kingdom \\
  $^{\mathsection}$ Brno University of Technology, Brno, Czech Republic
}
\email{florian.mai@idiap.ch}

\begin{document}
\ninept
\maketitle

\begin{abstract}
State-of-the-art ASR systems have achieved promising results by modeling local and global interactions separately. While the former can be computed efficiently, global interactions are usually modeled via attention mechanisms, which are expensive for long input sequences. Here, we address this by extending HyperMixer, an efficient alternative to attention exhibiting linear complexity, to the Conformer architecture for speech recognition, leading to HyperConformer. In particular, multi-head HyperConformer achieves comparable or higher recognition performance while being more efficient than Conformer in terms of inference speed, memory, parameter count, and available training data. HyperConformer achieves a word error rate of 2.9\% on LibriSpeech test-clean with less than 8M neural parameters and a peak memory during training of 5.7GB, hence trainable with accessible hardware. Encoder speed is between 38\% on mid-length speech and 56\% on long speech faster than an equivalent Conformer.\footnote{The HyperConformer recipe is publicly available in: \url{ https://github.com/speechbrain/speechbrain/tree/develop/recipes/LibriSpeech/ASR/transformer/}}
\end{abstract}

\noindent\textbf{Index Terms}: Hypernetworks, HyperMixer, Efficient Automatic Speech Recognition, LibriSpeech, SpeechBrain

\section{Introduction}
\label{sec:introduction}

Automatic Speech Recognition (ASR) technologies have greatly benefited from deep learning, reaching unprecedented levels of accuracy and pushing successful 
products to real-life use cases. Various architectures of ASR systems co-exist and deliver superlative performance depending on the task or domain of interest~\cite{nassif2019speech}. A prevalent family of ASR systems uses self-attention and Transformer neural networks to consume the input speech sequence and build powerful representations both at the acoustic and linguistic levels~\cite{karita2019comparative}. Indeed, the ability of Multi-Head Self-Attention (MHSA)~\cite{vaswani2017attention} to capture long-term dependencies via its sequence-long receptive field helped Transformer ASR architectures to outperform the previous state-of-the-art mostly composed with recurrent neural networks~\cite{karita2019comparative}. Nevertheless, ASR not only requires capturing global interactions describing the semantic and linguistic characteristics of the speech utterance but also modeling properly the local interactions that form the speech signal.

\begin{figure}[t]
    \centering
    \includegraphics[width=1\columnwidth]{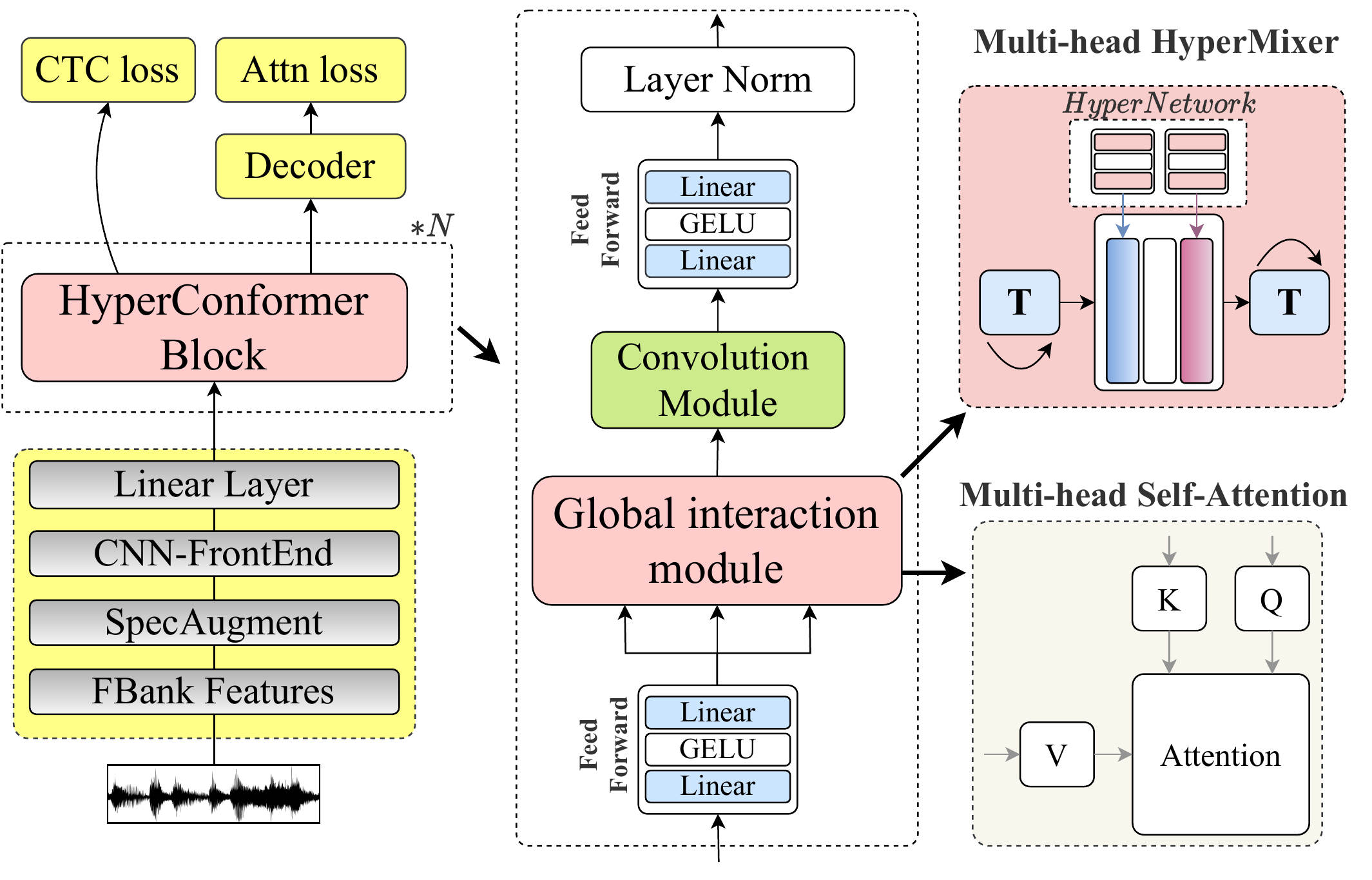}
    \caption{Layout of the general Conformer architecture. Global interactions can be modeled either with attention leading to a \emph{Conformer} or with HyperMixer to obtain \emph{HyperConformer}. \mbox{\textbf{\texttt{T}} represents} the transpose operation. Skip connections are omitted for simplicity. The global interaction module is combined sequentially with a convolution module to capture local dependencies, critical for speech-related tasks.}
    \label{fig:conformer-architecture}
\end{figure}

Conformer neural networks~\cite{Gulati2020-conformer} have been introduced to specifically address this issue. They combine Transformer and Convolutional Neural Network (CNN) blocks to capture the global and local dependencies respectively, leading to improved Word Error Rate (WER). Most prominently, variations of the Conformer, named Branchformer~\cite{branchformer} and E-Branchformer~\cite{e-branchformer} reached the lowest WER on the widely-adopted LibriSpeech dataset~\cite{panayotov2015librispeech} while being trained from scratch without external data. Following the local and global dependencies' assumption, Branchformer architecture physically create two branches per block (a dual path) in the architecture to capture independently and with adapted mechanisms (i.e., MHSA and CNN) both levels of dependencies. The latter branches are then merged and passed to the next architecture block. Such approaches are agnostic to the type of ASR decoding or processing, e.g., Transducers~\cite{graves2012sequence}, CTC only~\cite{ctc_loss}, or CTC and attention~\cite{watanabe2017hybrid}. However, they suffer from a major and well-documented efficiency issue as MHSA exhibits a quadratic complexity and memory time-dependency~\cite{mai2022hypermixer}. For instance, the MHSA block is among the most computationally demanding elements of any Transformer model. This is especially true for speech processing as input sequences are often long by nature e.g., longer than $30$ seconds for a few LibriSpeech utterances~\cite{gao22d_interspeech, parcollet21_interspeech}. In addition, large-scale and Transformer-based Self-Supervised Learning (SSL) models for speech recognition are commonly trained with sentences voluntarily cropped at $20$ to $25$ seconds. The latter transformation is necessary to enable training with top-tier GPU e.g., Tesla V100 or A100~\cite{babu2021xls}, also making it potentially intractable to train on more accessible compute infrastructures. This article focuses on retaining MHSA's global interactions capabilities beneficial to ASR while lowering significantly its computational and memory cost. 

% Paragraph on token mixing
How to efficiently compute interactions between tokens in Transformer-like architectures is an active area of research~\cite{tay2022efficient}.
Most works try to decrease the cost of attention directly, e.g., through a low-rank approximation~\cite{wang2020linformer}, linearization~\cite{katharopoulos2020transformers}, or the introduction of sparse attention patterns~\cite{child2019generating}.
However, token mixing can also be achieved from outside the framework of attention, opening up considerably novel opportunities for improvement. MLPMixer~\cite{tolstikhin2021mlp} was the first to learn a fixed-size MLP for modeling global interactions, with many to follow in the vision domain~\cite{liu2021pay, chen2021cyclemlp, wang2022dynamixer}. However, the fixed size hinders their adoption for domains with variable length signals. Existing approaches for speech have strong locality biases~\cite{speechmlp,sakuma2022mlp} and still rely on small attention modules for the best performance~\cite{sakuma2022mlp}.
Recently,~\cite{mai2022hypermixer} proposed HyperMixer for text processing, which achieved competitive performance to attention at a substantially lower cost in terms of computation and data.
Intuitively, HyperMixer constructs the token-mixing MLP of MLPMixer \emph{dynamically as a function of the data}, hence being amenable to variable length inputs.

This article introduces HyperConformer, a novel and simple-to-implement alternative to MHSA. It benefits from the linear time and memory complexity of HyperMixer while capturing both global and local dependencies from the speech signal necessary for ASR. The contributions are threefold. First, we formally describe HyperConformer and its main components (\S\ref{sec:hyperconformer}). Then, we introduce \emph{multi-head token mixing} to HyperMixer and HyperConformer to further improve the efficiency of both models. Finally, we open-source a training and inference ASR recipe within the widely adopted SpeechBrain toolkit~\cite{speechbrain}. Experiments are conducted in a relatively resource-constrained scenario with limited VRAM and neural parameters budgets to highlight the efficiency aspect of each evaluated model. Throughout the conducted ASR experiments on the LibriSpeech dataset (\S\ref{sec:exp-results}), HyperConformer consistently reaches the state-of-the-art Conformer baseline in terms of WER. In addition, HyperConformer shows between 37\% and 56\% reduction in processing time on mid-length and long speech, respectively. During training, it uses up to 30\% less memory, hence being trainable on GPUs from the Ti 70 family. Overall, HyperConformer offers a more accessible alternative to any ASR system previously based on Conformer models.

\section{HyperConformer}
\label{sec:hyperconformer}

Figure~\ref{fig:conformer-architecture} illustrates the different blocks of the introduced HyperConformer. It consists of four parts: Two feature mixing layers (feed-forward networks) at the bottom and top of the layer, a module for modeling local interactions, specifically the Convolution module introduced in~\cite{Gulati2020-conformer}, and a global interaction module. In the following, we discuss the global interaction modules bringing token mixing to the model. Other components of HyperConformer are identical to the Conformer~\cite{Gulati2020-conformer}.

\subsection{Capturing Global Interactions} 

Let $X \in \mathbb{R}^{N \times d}$ represent $N$ $d$-dimensional token vectors, also equivalent to a latent representation of speech coming from the previous layer on length $N$. The global interaction module $\operatorname{GI}: \mathbb{R}^{N \times d} \rightarrow \mathbb{R}^{N \times d}; X \mapsto X'$ is responsible for combining information from different tokens in such a way that every $X'_{:,j}$ contains information from every $X_{:,i}$. Such a behavior captures global interactions as it interconnects the different time steps of the given speech or latent sequence. This may be achieved, for instance, via multi-head attention or via HyperMixer.

\subsection{Multi-Head Self-Attention}
 At the core, Multi-Head Self-Attention (MHSA)~\cite{vaswani2017attention} relies on scaled dot-product attention:
\begin{equation*}
\text{Attention}(X) = \text{Softmax}(\frac{XX^T}{\sqrt{d_k}})X,    
\end{equation*}
which involves computing the dot product between every pair of input tokens, invoking memory and runtime complexity of $\mathcal{O}(N^2 \cdot d)$. The latter is responsible for the quadratic increase in memory and time consumption of standard Transformer architectures~\cite{mai2022hypermixer}. Further modeling capabilities are commonly obtained with the introduction of $k$ parallel heads, allowing the model to attend to information from different representation subspaces, i.e., different views of the data:
\begin{align*} 
\text{MHSA}(X) &= \text{Concat}(\text{head}_1,\dots,\text{head}_k)W^O, \\ 
\text{head}_i &= \text{Attention}(XW_i^Q,XW_i^K,XW_i^V),
\end{align*}
with $W^O, W_i^Q, W_i^K, W_i^V$ learnable weight parameters. 

\subsection{HyperMixer}

From a high-level perspective, HyperMixer achieves token mixing over variable length sequences by dynamically constructing a \emph{token mixing MLP} through the use of hypernetworks~\cite{ha2017hypernetworks}. The latter models specialize in generating neural network parameters, e.g., weights and biases. A token-mixing MLP is a multilayer perceptron $\tmmlp : \mathbb{R}^{d \times N} \rightarrow \mathbb{R}^{d \times N}$ that combines information from different tokens \emph{for each feature independently}, e.g., processing the Fbank coefficients of each time step of a sequence:

\begin{equation}
    \tmmlp(X)_{i,:} = \operatorname{LayerNorm}(W_1 (\sigma (W_2^{T} X_{i,:}^T))),\label{eq:mlp1}
\end{equation}
where $W_1, W_2 \in \mathbb{R}^{N \times d'}$ are weight matrices with the hidden layer size $d'$. $\sigma$ represents some non-linear activation function; we fix it to $\operatorname{GELU}$~\cite{hendrycks2016gaussian} following~\cite{mai2022hypermixer}.
Furthermore, we add layer normalization~\cite{ba2016layer} for improved stability.
Intuitively, the input layer $W_1$ decides to what degree each token's information should be sent to the hidden layer of $\tmmlp$, and the output layer $W_2$ decides for each token what information to extract from the hidden layer.

Importantly, $W_1, W_2$ themselves are not learnable parameters, which would require the input to be of the same fixed size at all times. Instead, $\operatorname{HyperMixer}(X; d, d')$, parameterized through the embedding dimension $d$ and the hidden layer size $d'$, first dynamically generates $W_1, W_2$ from the inputs themselves with the two hypernetworks $\operatorname{MLP^1}, \operatorname{MLP^2}$:

\begin{equation*}
    W_k(X) = \left(
\begin{array}{c}
\operatorname{MLP^k}(X_{:,1} + p_{:,1}) \\
\vdots \\
\operatorname{MLP^k}(X_{:,N} + p_{:,N})
\end{array}
\right) \in \mathbb{R}^{N \times d'}, k \in {1,2}.
\end{equation*}
$\operatorname{MLP^1}, \operatorname{MLP^2} : \mathbb{R}^{d} \rightarrow \mathbb{R}^{d'}$ contain the learnable parameters of HyperMixer, and $p_{:,j}$ are absolute position embeddings from standards Transformers~\cite{vaswani2017attention}. After generating the weights, Equation~\ref{eq:mlp1} is applied.
This determines the complexity of this model: $\mathcal{O}(N \cdot d \cdot d')$, which is the same asymptotic runtime as the feature mixing layers. Hence, HyperMixer turns the quadratic memory and inference time complexities to a linear regime.

\subsection{Multi-Head HyperMixer}

Analogously to MHSA, we propose an extension of HyperMixer to multi-head HyperMixer (MHHM) and HyperConformer, by introducing multiple token mixing heads. To this end, we create $k$ parallel $\operatorname{HyperMixer}^l( \cdot ; d / k, d' / k), l \in 0..k-1$, which each operates on $(d / k)$-dimensional feature subsets of $X$, whose outputs are again concatenated:
\begin{align*}
    \text{head}_l &= \operatorname{HyperMixer}^l (X_{:,(l\cdot(d/k)):(l+1\cdot(d/k))})\\
    \text{MHHM}(X) & = \text{Concat}(\text{head}_1,\dots,\text{head}_k)
\end{align*}
As a result, and conversely to MHSA, the runtime complexity even further reduces to $\mathcal{O}(k \cdot (N \cdot (d/k) \cdot (d'/k))) = \mathcal{O}(\frac{N \cdot d \cdot d'}{k})$.

\section{Experiments}
\label{sec:exp-results}

This section details the experimental setup (\ref{subsec:protocol}) used to evaluate HyperConformer against three baselines including the state-of-the-art Conformer. Models are compared both in terms of ASR performance (Section~\ref{subsec:results_asr}) and efficiency metrics (Section~\ref{subsec:results_analysis}).

\subsection{Experimental Setup}
\label{subsec:protocol}

Our experiments aim at assessing the effectiveness and efficiency of HyperConformer in comparison to Conformer. Hence, we compare vanilla Transformer~\cite{vaswani2017attention} and Conformer~\cite{Gulati2020-conformer} models to HyperMixer and HyperConformer. In practice, we swap the global interaction module, i.e., attention, from \texttt{regularMHA} (which uses absolute position embeddings~\cite{vaswani2017attention}) of Transformer and \texttt{RelPosMHAXL} (which uses relative position embeddings~\cite{dai-etal-2019-transformer}) of Conformer to our multi-head \texttt{HyperMixer} implementation. \\

\noindent\textbf{Datasets and Decoding.} We validate HyperConformer, on the LibriSpeech dataset~\cite{panayotov2015librispeech}. It is composed of $\sim$960h of transcribed speech in English. 
We perform ablations either training on the 100h set or the full, 960h set, and report results on the dev/test sets and clean/other partitions. Additionally, we use the text-only corpus for external language modeling (LM).\footnote{\raggedright Pre-trained LM from SpeechBrain available in: \scriptsize{\url{huggingface.co/speechbrain/asr-conformersmall-transformerlm-librispeech}}.} The LM is a Transformer based~\cite{vaswani2017attention} only-encoder model composed of 12 encoder layers, $d_{ffn}=3072$ and $d_{model}=768$, which accounts for 93.3M parameters. Word error rates are reported using beam search with and without LM shallow fusion. 

\noindent\textbf{Neural Architectures.} To gain a comprehensive understanding of performance and primary trade-offs, we ablate four different architectures in an encoder-decoder style: i) vanilla Transformer, ii) Conformer, iii) HyperMixer, and iv) HyperConformer. For the efficiency analysis only we also experiment with replacing \texttt{RelPosMHAXL} with \texttt{regularMHA} (Conformer-regular). All models use a 5K BPE sub-word unit~\cite{sennrich2016neural} vocabulary. This remains consistent across all experiments and models. At the bottom of the encoder, we incorporated a front-end module consisting of a 2-layer CNN that receives 80-dim log Mel filterbank features. We use SpecAugment~\cite{specaugment} during training with the default configuration in SpeechBrain. To correspond to accessible hardware as well as to emphasize low-compute resources performance, all models are conceived within a 25M parameter budget and trained with an 11GB memory constraint, corresponding to accessible GPU such as the Ti 80 family (or Ti 70 for the small version of HyperConformer). Hence, we select two model sizes for each architecture, i.e., 8 different scenarios. We use the same configuration, 10 encoder layers, and 8 attention or HyperMixing heads. However, we set $d_{model}=\{144,256\}$ for \texttt{\{base,medium\}} models, respectively. The feed-forward network dimensions is set to $d_{ffn}=4 \cdot d_{model}$ for all cases.
For simplicity, we set the hidden layer size $d'$ of $\tmmlp$ to $d' = d_{ffn}$.
We leave an exploration of this hyperparameter to future work.

\begin{table}[t]
    \centering
    \caption{Word error rates [\%] on the official LibriSpeech dev and test sets for models trained on 960h LibriSpeech set. The results include the four proposed encoder models, including our novel architecture, HyperConformer. We ablate two different model sizes for each architecture and list results with and without LM. The last column list the peak memory consumption [GB] of each architecture under the same training conditions. 
    }
    \label{tab:ls_full}
    \resizebox{\columnwidth}{!}{
    \begin{tabular}{c|c| cccc | cc | c}
    \toprule
    \textbf{Model} & \textbf{Par.} & \multicolumn{4}{c|}{\textbf{WER w/o LM}} & \multicolumn{2}{c|}{\textbf{WER w/ LM}} & \multicolumn{1}{c}{\textbf{Peak}} \\
     & & \multicolumn{2}{c}{\textbf{dev}} & \multicolumn{2}{c|}{\textbf{test}} & \multicolumn{2}{c|}{\textbf{test}} & \multicolumn{1}{c}{\textbf{Mem.}} \\
     \cmidrule(lr){3-4} \cmidrule(lr){5-6} \cmidrule(lr){7-8} 
     & \textit{[M]}& clean & other & clean & other & clean & other & \textit{[GB]}\\
     \midrule
    \rowcolor{Gray} \multicolumn{9}{c}{\textbf{Small sized models ($d_{model}=144$)}} \\
    \midrule
    Transformer & 6.1 & 7.7 & 15.6 & 7.8 & 15.8 & 3.9 & 8.2 & 6.45\\
    HyperMixer & 5.6 & 12.9 & 23.1 & 13.1 & 23.4 & 5.8 & 12.6 & 4.04 \\
    Conformer & 8.7 & 4.7 & 11.4 & 5.0 & 11.3 & 3.1 & 6.8 & 8.18 \\
    HyperConformer & 7.9 & 5.0 & 12.1 & 5.3 & 12.3 & 2.9 & 7.0 & 5.67 \\
     \midrule
     \rowcolor{Gray} \multicolumn{9}{c}{\textbf{Medium sized models ($d_{model}=256$)}} \\
     \midrule
    Transformer & 16.2 & 4.6 & 10.7 & 4.7 & 10.9 & 2.7 & 6.1 & 7.6 \\
    HyperMixer & 14.4 & 7.2 & 15.2 & 7.5 & 15.2 & 3.9 & 8.3 & 5.6 \\
    Conformer & 24.1 & 3.6 & 8.8 & 3.8 & 8.7 & 2.6 & 5.9 & 10.7 \\
    HyperConformer & 21.7 & 3.4 & 9.0 & 3.6 & 9.0 & 2.3 & 5.7 & 8.6 \\    
    \bottomrule
    \end{tabular}
    }
 \end{table}

\noindent\textbf{Training Hyperparameters.} Training is performed by combining the per-frame transformer decoder output probabilities and CTC~\cite{karita2019comparative}. The CTC loss~\cite{ctc_loss} is weighted by $\alpha=0.3$ during training. All the models use the same decoder, i.e., 4 Transformer layers. We follow the default training configuration of the LibriSpeech recipe from SpeechBrain.\footnote{Please refer to the SpeechBrain recipe located in \texttt{recipes/LibriSpeech/ASR/transformer}.} It uses Adam~\cite{kingma2014adam} optimizer, learning rate ($lr\!=1e^{-3}$) scheduler with warmup~\cite{vaswani2017attention} (25k steps warmup). We train for 110 epochs, i.e., $\sim$660k steps when full LibriSpeech and $\sim$70k when LibriSpeech 100h set. The recipe also uses dynamic batching, which reduces the overall training time. At decoding time, we use a beam size of 66 with a CTC weight of $ctc_{w}=0.4$. All of our experiments can be run on accessible GPUs starting from the Ti 70 family. 

\begin{figure*}[h!]
    \centering
    \begin{subfigure}{0.5\textwidth}
        \includegraphics[width=1\textwidth]{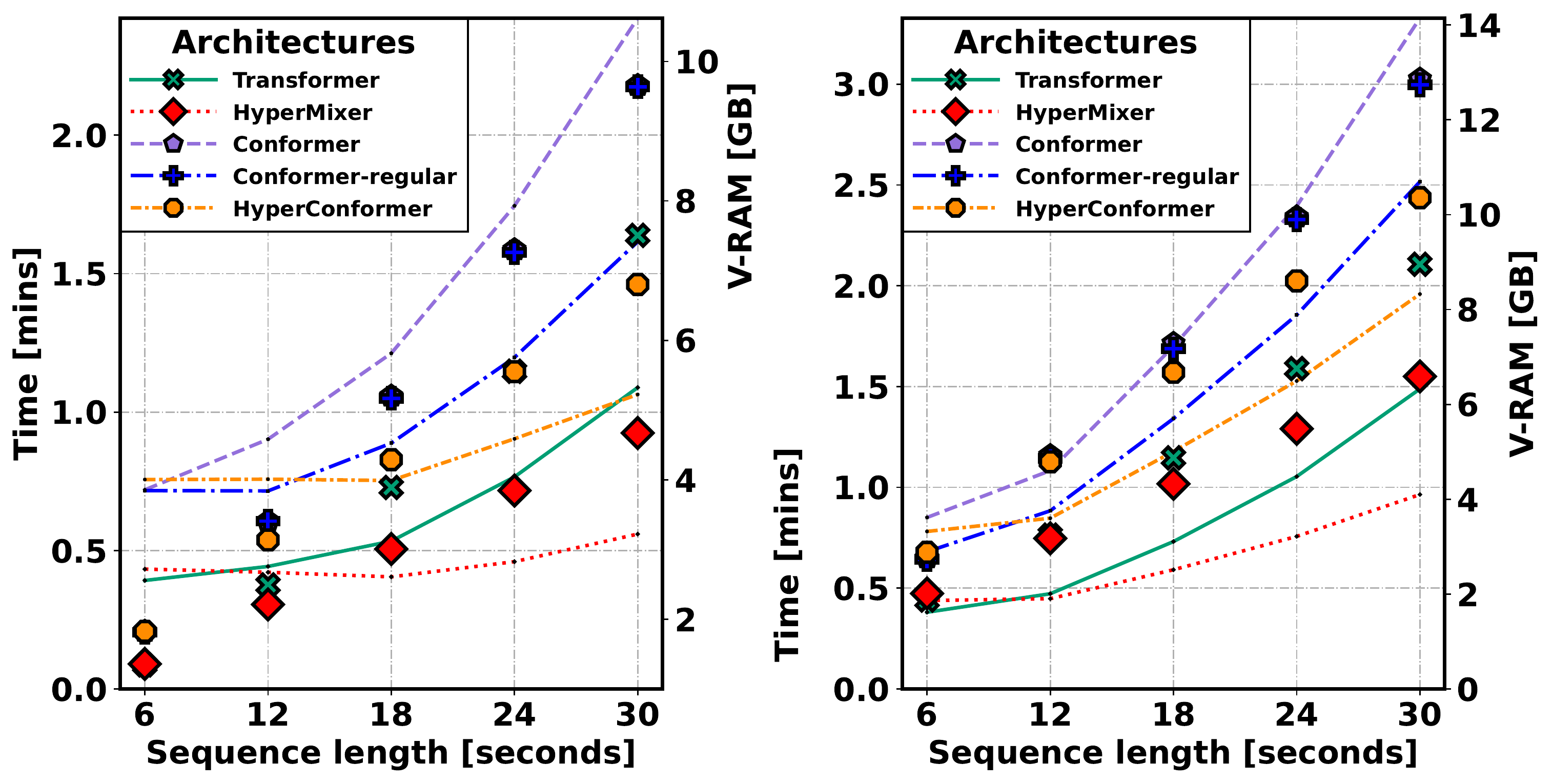}
        \caption{Forward pass of small (left) and medium sized (right) models. }
        \label{fig:comparison-of-models}
    \end{subfigure}%
    \begin{subfigure}{0.5\textwidth}
        \includegraphics[width=1\textwidth]{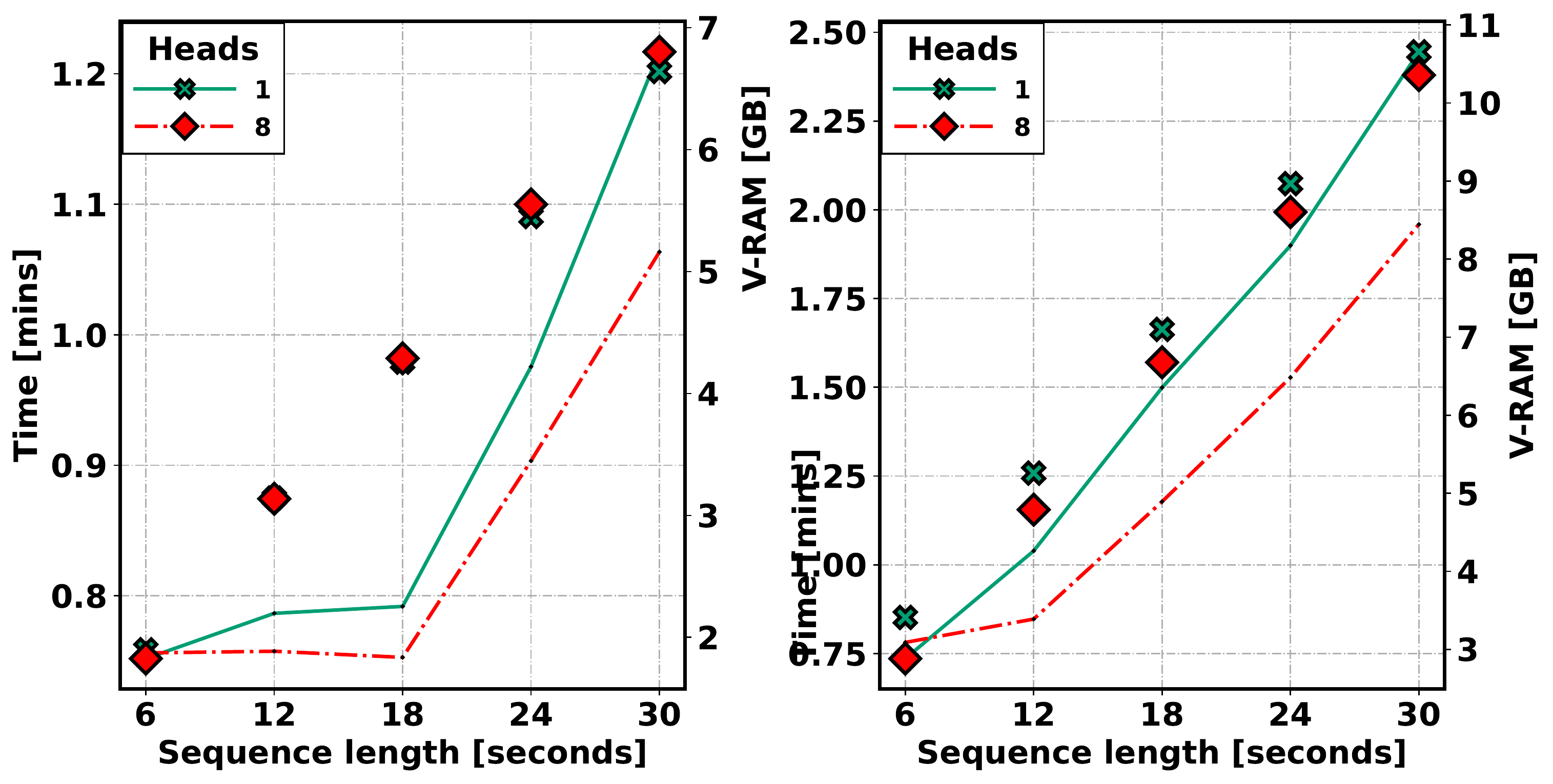}
        \caption{Forward pass of 1 and 8 heads for HyperConformer.}
        \label{fig:comparison-of-heads}
    \end{subfigure}%
    
    \caption{Overall time (minutes) and GPU consumption (GB) required by different architectures for sequences of different lengths. The left plot of (a) and (b) shows the small model, the right plot shows the mid-size model. Each sequence length in the x-axis represents 1000 samples from the LibriSpeech dataset. For all plots: Lines denotes time (left y-axis) and markers GPU consumption (right y-axis). Batch size is 16 for all configurations.}

    \label{fig:memory-by-seq-length}
\end{figure*}

\subsection{Results and Discussion}
Our experiments are designed to answer two questions: 1) Does HyperConformer perform competitive to Conformer in terms of word error rates? 2) Is HyperConformer more efficient than Conformer?

\subsubsection{Speech Recognition Results}
\label{subsec:results_asr}

We compare WERs of different state-of-the-art architectures for ASR, listing the results on Table~\ref{tab:ls_full}. We find that HyperMixer alone achieves acceptable performance, especially in combination with a language model, but trails behind Transformer and Conformer, in all cases.
We hypothesize that this is because the crucial local information in speech signals is difficult to pass through the hidden layer bottleneck of $\tmmlp$, which attention does not have. In contrast, HyperConformer performs comparable and often even better than Conformer in the medium-sized configuration.
For instance, \mbox{HyperConformer} beats Conformer by 0.17\% absolute WER on test-other with LM for the medium-sized model. 
We explain this as follows: In \mbox{HyperConformer}, i) the convolution module helps to model the local interactions between tokens, and ii) global interactions can be modeled in and passed through the multi-head HyperMixer's bottleneck effectively. 
Finally, we note that HyperConformer is amenable to scale, since moving from 7.9M $\rightarrow$ 21.7M, we obtain a 17.9\% relative reduction in WER on test-other with LM, similar to Conformer.

\subsubsection{Efficiency Analysis}
\label{subsec:results_analysis}

In~\cite{mai2022hypermixer} is shown that HyperMixer has efficiency benefits regarding processing speed and training data size. Here, we investigate if these properties also transfer to the speech domain, particularly, HyperConformer on the ASR task.

\noindent\textbf{Peak memory consumption}
The right hand side of Table~\ref{tab:ls_full} shows the peak memory consumption when training models of the same size on the same hardware. We observe that \mbox{HyperConformer} requires substantially less memory than Conformer (-30.6\% with small size and -19.7\% with medium size). The effect is stronger on small models than on large ones. Since larger models are wider (i.e., larger $d$ and $d'$), the feature mixing components as well as $\tmmlp$ require considerably more compute in comparison to attention, whose complexity depends primarily on the sequence length, which remains the same between training scenarios.

\noindent\textbf{Resource consumption depending on sequence length}
The main advantage of HyperMixer is its linear complexity compared to attention's quadratic complexity.
To investigate this property, we measure the peak memory and processing time of the encoder as a function of the length of the speech sample. To this end, we synthesize 1,000 sentences of 6, 12, 18, 24, and 30 seconds each by concatenating multiple signals from the LibriSpeech dataset. Figure~\ref{fig:comparison-of-models} shows the resource consumption of all models. While HyperConformer and Conformer require similar processing time at short sequences, HyperConformer is considerably faster at mid-length (18s, small: 37.9\%, mid: 15.2\%) and long sequences (30s, small 56.1\%, mid: 34.2\%), demonstrating its better asymptotic complexity compared to Conformer. Note that  Conformer with \texttt{regularMHA} is more efficient than \texttt{RelPosMHAXL}. However, this would lead to a performance loss~\cite{Gulati2020-conformer}, and HyperConformer is still substantially more efficient.

\noindent\textbf{Number of heads}
An important technical novelties is the introduction of multi-head HyperMixer, which allows for multiple parallel views on the data analogous to multi-head attention, while at the same time reducing the model's complexity.
In preliminary experiments, we found that \mbox{HyperConformer} with $k = 8$ heads performs as well as with $k = 1$ head. At the same time, moving from a single head to 8 heads reduces the number of parameters in the model by 7.1\% in the small model and 20.8\% in the mid-size model.
Moreover, as Figure~\ref{fig:comparison-of-heads} shows, the processing time is reduced substantially by up to 12.6\% (small) and 19.9\% (mid-size) on the longest sequences.

\noindent\textbf{Low-resource scenario}
HyperMixer is reported to work better than MHSA in the low-resource scenario~\cite{mai2022hypermixer}. Here, we conduct an initial experiment to test whether \mbox{HyperConformer} inhibits the same characteristic. To this end, we compare \mbox{HyperConformer} to Conformer on the 100h LibriSpeech subset, which is 10 times smaller than the full dataset.
All other training parameters remain the same. Table~\ref{tab:low-resource} shows the results.
In this scenario, \mbox{HyperConformer} performs around 20\% better than Conformer, suggesting better data efficiency.

\begin{table}[h]
    \centering
    \caption{Performance of Conformer and HyperConformer when trained on 100h LibriSpeech (10$\times$ less data). Percentage in brackets shows relative WER reduction on test-other with LM.}
    \label{tab:low-resource}
    \begin{tabular}{c|l|l}
    \toprule
    \textbf{Model} & \textbf{Small size} & \textbf{Medium size} \\
    \midrule
        Conformer & 8.29  & 7.57 \\
        HyperConformer & 6.76 (-18.5\%) & 5.80 (-23.4\%) \\
    \bottomrule
    \end{tabular}
\end{table}

\section{Conclusion}

\mbox{HyperConformer} is a new architecture for efficient ASR introduced in this work. It integrates the benefits of the Convolution module from Conformer, which models local interactions, and the hypernetwork-based architecture, HyperMixer, which models global interactions. We were able to attain comparable or lower WERs (2.28/5.42 in test clean/other) \mbox{HyperConformer} when compared to Conformer. In addition, this novel architecture is substantially faster on long sequences, while also requiring less GPU memory during training. We believe \mbox{HyperConformer} is a green alternative to previous established Transformer and Conformer based models for ASR.

\bibliographystyle{IEEEtran}
\bibliography{biblio}

\begin{thebibliography}{10}
\providecommand{\url}[1]{#1}
\csname url@samestyle\endcsname
\providecommand{\newblock}{\relax}
\providecommand{\bibinfo}[2]{#2}
\providecommand{\BIBentrySTDinterwordspacing}{\spaceskip=0pt\relax}
\providecommand{\BIBentryALTinterwordstretchfactor}{4}
\providecommand{\BIBentryALTinterwordspacing}{\spaceskip=\fontdimen2\font plus
\BIBentryALTinterwordstretchfactor\fontdimen3\font minus
  \fontdimen4\font\relax}
\providecommand{\BIBforeignlanguage}[2]{{%
\expandafter\ifx\csname l@#1\endcsname\relax
\typeout{** WARNING: IEEEtran.bst: No hyphenation pattern has been}%
\typeout{** loaded for the language `#1'. Using the pattern for}%
\typeout{** the default language instead.}%
\else
\language=\csname l@#1\endcsname
\fi
#2}}
\providecommand{\BIBdecl}{\relax}
\BIBdecl

\bibitem{nassif2019speech}
A.~B. Nassif, I.~Shahin, I.~Attili, M.~Azzeh, and K.~Shaalan, ``Speech
  recognition using deep neural networks: A systematic review,'' \emph{IEEE
  access}, vol.~7, pp. 19\,143--19\,165, 2019.

\bibitem{karita2019comparative}
S.~Karita, N.~Chen, T.~Hayashi, T.~Hori, H.~Inaguma, Z.~Jiang, M.~Someki,
  N.~E.~Y. Soplin, R.~Yamamoto, X.~Wang \emph{et~al.}, ``A comparative study on
  transformer vs rnn in speech applications,'' in \emph{2019 IEEE Automatic
  Speech Recognition and Understanding Workshop (ASRU)}.\hskip 1em plus 0.5em
  minus 0.4em\relax IEEE, 2019, pp. 449--456.

\bibitem{vaswani2017attention}
A.~Vaswani, N.~Shazeer, N.~Parmar, J.~Uszkoreit, L.~Jones, A.~N. Gomez,
  {\L}.~Kaiser, and I.~Polosukhin, ``Attention is all you need,''
  \emph{Advances in neural information processing systems}, vol.~30, 2017.

\bibitem{Gulati2020-conformer}
\BIBentryALTinterwordspacing
A.~Gulati, J.~Qin, C.-C. Chiu, N.~Parmar, Y.~Zhang, J.~Yu, W.~Han, S.~Wang,
  Z.~Zhang, Y.~Wu, and R.~Pang, ``{Conformer: Convolution-augmented Transformer
  for Speech Recognition},'' in \emph{Proc. Interspeech 2020}, 2020, pp.
  5036--5040. [Online]. Available:
  \url{http://dx.doi.org/10.21437/Interspeech.2020-3015}
\BIBentrySTDinterwordspacing

\bibitem{branchformer}
\BIBentryALTinterwordspacing
Y.~Peng, S.~Dalmia, I.~Lane, and S.~Watanabe, ``Branchformer: Parallel
  {MLP}-attention architectures to capture local and global context for speech
  recognition and understanding,'' in \emph{Proceedings of the 39th
  International Conference on Machine Learning}, ser. Proceedings of Machine
  Learning Research, K.~Chaudhuri, S.~Jegelka, L.~Song, C.~Szepesvari, G.~Niu,
  and S.~Sabato, Eds., vol. 162.\hskip 1em plus 0.5em minus 0.4em\relax PMLR,
  17--23 Jul 2022, pp. 17\,627--17\,643. [Online]. Available:
  \url{https://proceedings.mlr.press/v162/peng22a.html}
\BIBentrySTDinterwordspacing

\bibitem{e-branchformer}
K.~Kim, F.~Wu, Y.~Peng, J.~Pan, P.~Sridhar, K.~J. Han, and S.~Watanabe,
  ``E-branchformer: Branchformer with enhanced merging for speech
  recognition,'' in \emph{2022 IEEE Spoken Language Technology Workshop
  (SLT)}.\hskip 1em plus 0.5em minus 0.4em\relax IEEE, 2023, pp. 84--91.

\bibitem{panayotov2015librispeech}
V.~Panayotov, G.~Chen, D.~Povey, and S.~Khudanpur, ``Librispeech: an asr corpus
  based on public domain audio books,'' in \emph{2015 IEEE international
  conference on acoustics, speech and signal processing (ICASSP)}.\hskip 1em
  plus 0.5em minus 0.4em\relax IEEE, 2015, pp. 5206--5210.

\bibitem{graves2012sequence}
A.~Graves, ``Sequence transduction with recurrent neural networks,''
  \emph{arXiv preprint arXiv:1211.3711}, 2012.

\bibitem{ctc_loss}
A.~Graves and A.~Graves, ``Connectionist temporal classification,''
  \emph{Supervised sequence labelling with recurrent neural networks}, pp.
  61--93, 2012.

\bibitem{watanabe2017hybrid}
S.~Watanabe, T.~Hori, S.~Kim, J.~R. Hershey, and T.~Hayashi, ``Hybrid
  ctc/attention architecture for end-to-end speech recognition,'' \emph{IEEE
  Journal of Selected Topics in Signal Processing}, vol.~11, no.~8, pp.
  1240--1253, 2017.

\bibitem{mai2022hypermixer}
F.~Mai, A.~Pannatier, F.~Fehr, H.~Chen, F.~Marelli, F.~Fleuret, and
  J.~Henderson, ``Hypermixer: An mlp-based low cost alternative to
  transformers,'' in \emph{Proceedings of the 61st Annual Meeting of the
  Association for Computational Linguistics (Volume 1: Long Papers)}, 2023.

\bibitem{gao22d_interspeech}
Y.~Gao, J.~Fernandez-Marques, T.~Parcollet, A.~Mehrotra, and N.~Lane,
  ``{Federated Self-supervised Speech Representations: Are We There Yet?}'' in
  \emph{Proc. Interspeech 2022}, 2022, pp. 3809--3813.

\bibitem{parcollet21_interspeech}
T.~Parcollet and M.~Ravanelli, ``{The Energy and Carbon Footprint of Training
  End-to-End Speech Recognizers},'' in \emph{Proc. Interspeech 2021}, 2021, pp.
  4583--4587.

\bibitem{babu2021xls}
A.~Babu, C.~Wang, A.~Tjandra, K.~Lakhotia, Q.~Xu, N.~Goyal, K.~Singh, P.~von
  Platen, Y.~Saraf, J.~Pino \emph{et~al.}, ``Xls-r: Self-supervised
  cross-lingual speech representation learning at scale,'' \emph{arXiv preprint
  arXiv:2111.09296}, 2021.

\bibitem{tay2022efficient}
Y.~Tay, M.~Dehghani, D.~Bahri, and D.~Metzler, ``Efficient transformers: A
  survey,'' \emph{ACM Computing Surveys}, vol.~55, no.~6, pp. 1--28, 2022.

\bibitem{wang2020linformer}
S.~Wang, B.~Z. Li, M.~Khabsa, H.~Fang, and H.~Ma, ``Linformer: Self-attention
  with linear complexity,'' \emph{arXiv preprint arXiv:2006.04768}, 2020.

\bibitem{katharopoulos2020transformers}
A.~Katharopoulos, A.~Vyas, N.~Pappas, and F.~Fleuret, ``Transformers are rnns:
  Fast autoregressive transformers with linear attention,'' in
  \emph{International Conference on Machine Learning}.\hskip 1em plus 0.5em
  minus 0.4em\relax PMLR, 2020, pp. 5156--5165.

\bibitem{child2019generating}
R.~Child, S.~Gray, A.~Radford, and I.~Sutskever, ``Generating long sequences
  with sparse transformers,'' \emph{arXiv preprint arXiv:1904.10509}, 2019.

\bibitem{tolstikhin2021mlp}
I.~O. Tolstikhin, N.~Houlsby, A.~Kolesnikov, L.~Beyer, X.~Zhai, T.~Unterthiner,
  J.~Yung, A.~Steiner, D.~Keysers, J.~Uszkoreit \emph{et~al.}, ``Mlp-mixer: An
  all-mlp architecture for vision,'' \emph{Advances in neural information
  processing systems}, vol.~34, pp. 24\,261--24\,272, 2021.

\bibitem{liu2021pay}
H.~Liu, Z.~Dai, D.~So, and Q.~V. Le, ``Pay attention to mlps,'' \emph{Advances
  in Neural Information Processing Systems}, vol.~34, pp. 9204--9215, 2021.

\bibitem{chen2021cyclemlp}
S.~Chen, E.~Xie, C.~Ge, D.~Liang, and P.~Luo, ``Cyclemlp: A mlp-like
  architecture for dense prediction,'' \emph{arXiv preprint arXiv:2107.10224},
  2021.

\bibitem{wang2022dynamixer}
Z.~Wang, W.~Jiang, Y.~M. Zhu, L.~Yuan, Y.~Song, and W.~Liu, ``Dynamixer: a
  vision mlp architecture with dynamic mixing,'' in \emph{International
  Conference on Machine Learning}.\hskip 1em plus 0.5em minus 0.4em\relax PMLR,
  2022, pp. 22\,691--22\,701.

\bibitem{speechmlp}
\BIBentryALTinterwordspacing
C.~Xing, D.~Wang, L.~Dai, Q.~Liu, and A.~Avila, ``Speech-{MLP}: a simple {MLP}
  architecture for speech processing,'' 2022. [Online]. Available:
  \url{https://openreview.net/forum?id=-u8EliRNW8k}
\BIBentrySTDinterwordspacing

\bibitem{sakuma2022mlp}
J.~Sakuma, T.~Komatsu, and R.~Scheibler, ``Mlp-asr: Sequence-length agnostic
  all-mlp architectures for speech recognition,'' \emph{arXiv preprint
  arXiv:2202.08456}, 2022.

\bibitem{speechbrain}
M.~Ravanelli, T.~Parcollet, P.~Plantinga, A.~Rouhe, S.~Cornell, L.~Lugosch,
  C.~Subakan, N.~Dawalatabad, A.~Heba, J.~Zhong \emph{et~al.}, ``Speechbrain: A
  general-purpose speech toolkit,'' \emph{arXiv preprint arXiv:2106.04624},
  2021.

\bibitem{ha2017hypernetworks}
\BIBentryALTinterwordspacing
D.~Ha, A.~M. Dai, and Q.~V. Le, ``Hypernetworks,'' in \emph{International
  Conference on Learning Representations}, 2017. [Online]. Available:
  \url{https://openreview.net/forum?id=rkpACe1lx}
\BIBentrySTDinterwordspacing

\bibitem{hendrycks2016gaussian}
D.~Hendrycks and K.~Gimpel, ``Gaussian error linear units (gelus),''
  \emph{arXiv preprint arXiv:1606.08415}, 2016.

\bibitem{ba2016layer}
J.~L. Ba, J.~R. Kiros, and G.~E. Hinton, ``Layer normalization,'' \emph{arXiv
  preprint arXiv:1607.06450}, 2016.

\bibitem{dai-etal-2019-transformer}
\BIBentryALTinterwordspacing
Z.~Dai, Z.~Yang, Y.~Yang, J.~Carbonell, Q.~Le, and R.~Salakhutdinov,
  ``Transformer-{XL}: Attentive language models beyond a fixed-length
  context,'' in \emph{Proceedings of the 57th Annual Meeting of the Association
  for Computational Linguistics}.\hskip 1em plus 0.5em minus 0.4em\relax
  Florence, Italy: Association for Computational Linguistics, Jul. 2019, pp.
  2978--2988. [Online]. Available: \url{https://aclanthology.org/P19-1285}
\BIBentrySTDinterwordspacing

\bibitem{sennrich2016neural}
R.~Sennrich, B.~Haddow, and A.~Birch, ``Neural machine translation of rare
  words with subword units,'' in \emph{Proceedings of the 54th Annual Meeting
  of the Association for Computational Linguistics}, 2016, pp. 1715--1725.

\bibitem{specaugment}
\BIBentryALTinterwordspacing
D.~S. Park, W.~Chan, Y.~Zhang, C.-C. Chiu, B.~Zoph, E.~D. Cubuk, and Q.~V. Le,
  ``{SpecAugment: A Simple Data Augmentation Method for Automatic Speech
  Recognition},'' in \emph{Proc. Interspeech 2019}, 2019, pp. 2613--2617.
  [Online]. Available: \url{http://dx.doi.org/10.21437/Interspeech.2019-2680}
\BIBentrySTDinterwordspacing

\bibitem{kingma2014adam}
D.~P. Kingma and J.~Ba, ``Adam: A method for stochastic optimization,''
  \emph{arXiv preprint arXiv:1412.6980}, 2014.

\end{thebibliography}

\end{document}